\pdfoutput=1

\documentclass[journal]{IEEEtran}
\usepackage{times}

\usepackage{algorithm,algpseudocode,algorithmicx}
\usepackage{soul}
\usepackage{url}
\usepackage[hidelinks]{hyperref}
\usepackage[utf8]{inputenc}
\usepackage[small]{caption}
\usepackage{graphicx}
\usepackage[graphicx]{realboxes}
\graphicspath{{Figures/}}
\usepackage{epstopdf}
\usepackage{amsmath}
\usepackage{amsfonts}
\usepackage{booktabs}
\usepackage{color}
\usepackage{makeidx}
\usepackage{multirow}
\ifCLASSOPTIONcompsoc
\usepackage[nocompress]{cite}
\else
\usepackage{cite}
\fi

\ifCLASSINFOpdf
\else
\fi
\begin{document}
%

\title{Bi-fidelity Evolutionary Multiobjective Search for Adversarially Robust Deep Neural Architectures 
}
%
%
%

\author{Jia~Liu,~\IEEEmembership{~Student Member,~IEEE}
        and~Ran~Cheng,~\IEEEmembership{~Senior Member,~IEEE}
        and~Yaochu~Jin,~\IEEEmembership{~Fellow,~IEEE}
\thanks{Note: A preliminary version of this work was presented at the International Workshop on Trustworthy Federated Learning in Conjunction with IJCAI 2022 (FL-IJCAI'22), August 3, 2022, Vienna, Austria.}
}

%
%

\markboth{Journal of \LaTeX\ Class Files,~Vol.~xx, No.xx, 2022}%
{Shell \MakeLowercase{\textit{et al.}}: Bare Demo of IEEEtran.cls for IEEE Journals}
%



\maketitle

\begin{abstract}
Deep neural networks have been found vulnerable to adversarial attacks, thus raising potentially concerns in security-sensitive contexts. To address this problem, recent research has investigated the adversarial robustness of deep neural networks from the architectural point of view.
However, searching for architectures of deep neural networks is computationally expensive, particularly when coupled with adversarial training process. To meet the above challenge, this paper proposes a bi-fidelity multiobjective neural architecture search approach. First, we formulate the NAS problem for enhancing adversarial robustness of deep neural networks into a multiobjective optimization problem.
Specifically, in addition to a low-fidelity performance predictor as the first objective, we leverage an auxiliary-objective -- the value of which is the output of a surrogate model trained with high-fidelity evaluations.
Secondly, we reduce the computational cost by combining three performance estimation methods, i.e., parameter sharing, low-fidelity evaluation, and surrogate-based predictor. The effectiveness of the proposed approach is confirmed by extensive experiments conducted on CIFAR-10, CIFAR-100 and SVHN datasets.
\end{abstract}

\begin{IEEEkeywords}
Adversarial attacks, neural architecture search, surrogate, low-fidelity, multiobjectivization.
\end{IEEEkeywords}

%
\IEEEpeerreviewmaketitle

\section{Introduction} \label{secI}
%
%
%
%
\IEEEPARstart{D}{eep} networks (DNNs) have been successfully employed to perform various complex applications in recent years, e.g., image classification \cite{krizhevsky2012imagenet,zeiler2014visualizing,lin2013network,simonyan2014very,szegedy2015going,he2016deep,huang2017densely,sabour2017dynamic}, object detection \cite{ren2015faster,ren2016object}, and language modeling \cite{sutskever2014sequence}.
However, DNN models are vulnerable to adversarial examples that are intentionally crafted with imperceptible perturbations \cite{szegedy2013intriguing}.
With the expanding use of DNNs on safety-sensitive applications such as self-driving cars and facial recognition systems, it is essential to develop accurate and robust models.
Much effort has been made to tackle the threat of adversarial examples.
The defense techniques can be divided into two types \cite{yuan2019adversarial}, namely proactive techniques that strengthen DNNs before meeting adversarial examples, including adversarial training, defensive distillation, and classifier robustifying, and reactive techniques that detect adversarial examples after the construction of DNNs, including adversary detection, input reconstruction, and network verification.

Despite the considerable effort on defense strategies, a majority of researchers carried out experiments on the basis of one or two specific manually designed convolutional neural networks (CNNs) such as PreAct ResNet \cite{he2016identity}, and WideResNet \cite{zagoruyko2016wide}.
Recently, neural architecture search (NAS) has attracted increasing attention and achieved promising performance on a variety of tasks.
Early NAS algorithms \cite{zoph2016neural} suffer from an extremely heavy computational burden since evaluating the ground-truth performance of each candidate architecture requires training the network from scratch.
Thus, how quickly and accurately estimate the performance of candidate architecture is a major challenge in NAS.
Various estimation strategies have been employed to reduce the computational cost. Some methods proposed to build low-fidelity proxy networks with fewer layers or fewer channels \cite{pham2018efficient, real2019regularized, wu2019fbnet} and train them to solve proxy tasks of smaller scales \cite{cai2018efficient, wu2019fbnet, jia2021multi}.
However, it has been found that the architectures obtained from such proxy tasks may not perform well on the target task.
To address this issue, parameter sharing \cite{pham2018efficient, cai2019once} and predictor-based evaluators \cite{liu2018progressive, sun2019surrogate} were proposed as another two efficient methods to estimate the performance of architectures.
The parameter-sharing method trains a supernet with all the attainable structures as its subnets.
During the search, the parameters of the subnets are derived from the weights inherited from the supernet.
However, the weights derived from the supernet may not be entirely reliable for assessing the performance of candidate architectures.
Hence, it has been proposed to fine-tune the shared weights using a few gradient steps \cite{chen2020multi}.
The predictor-based evaluators were proposed to accelerate NAS by training a surrogate model using a set of DNNs labeled with their corresponding performance.
During the search process, the surrogate model directly predicts the performance of the newly searched architectures.
To further accelerate the NAS process, Luo \textit{et al}. \cite{luo2018neural} attempted to combine parameter-sharing evaluator and predictor-based evaluator.
However, they failed to find architectures with better performance compared with using expensive evaluation.

Despite the remarkable progress introduced above, existing NAS methods mainly focus on improving classification accuracy and are limited by intensive computation and memory costs.
Only a few studies have attempted to understand adversarial robustness from an architectural point of view.
Alparslan and Kim \cite{alparslan2021atras} explored how the architecture size impacts the model robustness. Huang \textit{et al.} \cite{huang2021exploring} used grid search to investigate the impact of model width and depth on adversarial trained WideResNet-34-10 \cite{zagoruyko2016wide}. 
However, the above research considers model parameters instead of the architecture topology. 
Robust architecture search (RAS) \cite{vargas2019evolving} adopted an evolutionary algorithm to search for robust architectures that are potentially less sensitive to transferable black-box attacks from a broader search space.
The performance of the models discovered by RAS is measured by the validation accuracy on clean images plus the accuracy on 2812 adversarial examples.
Although the evolved architectures were less sensitive in this scenario, the generality is limited by the pre-labeled adversarial examples.
Dong \textit{et al.} \cite{dong2019neural} employed a differentiable NAS framework for adversarial medical image segmentation, which automatically find the architecture of a discriminator.
In \cite{guo2020meets}, RobNet employed adversarial training and one-shot NAS to understand the performance of architectures against adversarial attacks, showing that densely connected structures contribute a lot to the model robustness.
Yue \textit{et al}. \cite{yue2020effective} proposed an algorithm called E2RNAS, which combines the multiple-gradient descent algorithm with the bi-level optimization to search for architectures that are effective, efficient, and robust.
It was the first work proposed to simultaneously optimize the performance, robustness, and resource constraint.
Despite that the search cost of E2RNAS is less than one GPU day, the classification accuracies (6\%$\sim$11\%) of the obtained architecture under PGD attack are unsatisfactory.
To search for robust architectures at targeted capacities, Ning \textit{et al}. \cite{ning2020discovering} proposed multi-shot NAS which trains eight supernets with different initial channel numbers.
The discovered architectures outperform peer competitors at targeted capacities, but the supernet training cost is much higher than one-shot NAS.
Devaguptapu \textit{et al}. \cite{devaguptapu2021adversarial} analyzed the adversarial robustness of manually designed CNNs as well as NAS-based models.
Since models in \cite{devaguptapu2021adversarial} were trained without any defense techniques, few models can resist strong adversarial attacks.
Chen \textit{et al.} \cite{chen2020anti} proposed ABanditNAS that employs an anti-bandit strategy to search multiple cells composed of denoising blocks, weight-free operations, Gabor filters, and convolutions.
Cazenavette \textit{et al.} \cite{cazenavette2021architectural} proposed a deep pursuit algorithm that formulates the architecture search as a global sparse coding problem that jointly computes all network activations.
Hosseini \textit{et al.} \cite{Hosseini_2021_CVPR} introduced DSRNA, which employs differentiable NAS to maximize the two proposed robustness metrics, i.e., certified lower bound and Jacobian norm bound.
Wang \textit{et al.} \cite{wang2021neural} proposed a multiobjective gradient optimization method and designed a new search space to automatically search for robust architectures for AI-enabled Internet-of-Things (AIoT) systems.
Mok \textit{et al.} \cite{mok2021advrush} proposed AdvRush that employs DARTS \cite{liu2018darts} as the backbone search algorithm to search for architectures with a smooth input loss landscape.
Focused on the robustness of tiny networks, TAM-NAS \cite{xie2021tiny} leveraged one-shot NAS and multiobjective optimization to obtain trade-off solutions between the adversarial accuracy, the clean accuracy, and the mode size. However, the work mainly focused on tiny neural networks.
In our previous work, MORAS \cite{jia2021multi}, we introduced a robustness measure against multiple adversarial attacks as an objective function in addition to the performance on the original dataset to search for architectures that are less sensitive to various adversarial attacks.
However, the search process is time-consuming since each architecture in the population has to be trained to obtain the fitness values.

To further improve the efficiency of NAS for improving adversarial robustness of DNNs, in this work, we propose the MORAS-S method -- \textbf{M}ulti-\textbf{O}bjective \textbf{R}obust \textbf{A}rchitecture \textbf{S}earch based on a \textbf{S}urrogate as a \textbf{H}elper-objective.
Specifically, the main contributions of this work are as follows:
\begin{itemize}
	\item To enhance the adversarial robustness of DNNs, we formulate the NAS problem into a multiobjective optimization problem by introducing an online surrogate model as an additional objective -- the auxiliary-objective -- to predict the high-fidelity fitness values of candidate architectures.
	\item 
	To accelerate the search process, we predict the performance of candidate architectures by combining parameter sharing with a predictor-based evaluator, where the parameters directly inherited from robust supernet training will be used as the performance evaluator. One one hand, the performance calculated from a partial validation set is used as a low-fidelity fitness evaluation. One the other hand, we calculate the performance of architecture on the entire validation set as the high-fidelity fitness evaluation, and a surrogate model built from the high-fidelity fitness evaluation will be used to approximate the high-fidelity fitness function.
	\item Experiments on benchmark datasets demonstrate that the proposed MORAS-SH method efficiently finds robust architectures with comparable classification accuracy.
\end{itemize}

The remainder of this paper is organized as follows. The next section will briefly describe the related work. Section \ref{iii} elaborates the proposed approach. Experimental settings and implementation details are presented in Section \ref{iv}, followed by descriptions of the experimental results and discussion in Section \ref{v}. Finally, we summarize our findings along with our future work.

\section{Related Work} \label{ii}
Research has made promising progress over the past few years to obtain more robust networks efficiently. In the remainder of this section, we will review some of the approaches relevant to this work, including adversarial attacks and defenses, methods to accelerate network performance assessment, and evolutionary multiobjective NAS.
\subsection{Adversarial Attack and Defense} \label{pgd}
Deep learning models can be misled by adversarial attacks, such as fast gradient sign method (FGSM) \cite{goodfellow2014explaining}, basic iterative method \cite{kurakin2016adversarial}, and C\&W attack \cite{carlini2017towards}.
One of the strongest adversarial attacks, PGD \cite{madry2018towards}, which combines randomized initialization with multi-step attacks, can be expressed as follows:
\begin{eqnarray}
	{\mathbf{X}^*_{0}} &=& \mathbf{X} + \mathcal{U}(-\epsilon, \epsilon), \\
	{\mathbf{X}^*_{n+1}} &=& \Pi_{\mathbf{X},\epsilon}\lbrace{\mathbf{X}^*_{n} + \alpha \cdot \textrm{sign}({\nabla}_{\mathbf{X}^*_{n}}{J}(\mathbf{X}^*_{n},y))}\rbrace
	\label{2.3}
\end{eqnarray}
where $X^*$ denotes the adversarial examples, $X$ denotes the original examples, $\mathcal{U}$ refers to a uniform distribution, $\epsilon$ is a hyper-parameter that controls the magnitude of the disturbance, $\alpha$ represents the step size, ${\nabla}_{\mathbf{X}^*_{n}}$ measures the gradient of the loss function $J$ with respect to $\mathbf{X}^*_{n}$, $y$ denotes the true label, and $\Pi_{\mathbf{X},\epsilon}(B)$ denotes the projection to $B(\mathbf{X}, \epsilon)$.

Extensive counter-measures have been designed to improve the robustness of deep learning models, such as adversarial training \cite{goodfellow2014explaining}, defensive distillation \cite{papernot2016distillation}, feature squeezing \cite{xu2017feature}, defense-GAN \cite{samangouei2018defense}, and autoencoder-based denoising \cite{liao2018defense}. Among those, adversarial training \cite{goodfellow2014explaining, madry2018towards}, turning out to be an effective method for improving robustness, improves the robustness of a network by training it together with adversarial samples.
Adversarial training works by minimizing the weighted training loss on clean and adversarial examples.
In this work, we employ PGD adversarial training (PGD-AT) to train the supernet and the architectures that are going to be evaluated for final evaluation. The PGD-AT process can be mathematically expressed as:
\begin{equation}
	\begin{split}
		\mathop {\min}\limits_{\theta} \mathbb{E}_{(\mathbf{X},y)\sim D} {J}(PGD(\mathbf{X}, \Omega), y) \\
	\end{split}
\end{equation}
where $\Omega$ is the threat model, $\theta$ is the model parameters.
\subsection{Neural Architecture Evaluators}
In NAS, if the network is completely trained from scratch, it often takes a lot of time to evaluate the performance of a network. In practice, parameter sharing and predictor-based evaluators are two commonly used techniques to efficiently evaluate architectures without training each candidate architecture from scratch.

\textbf{Parameter Sharing.}
Parameter sharing \cite{pham2018efficient}, also known as weight sharing, is the process of building and training a super-large network within a given search space, and then the subnet directly shares the parameters from supernet.
This over-parameterized supernet will contain all possible subnets.
Therefore, the evaluation of subnets greatly reduces the time to evaluate candidate architectures because they share the parameters of the supernet without training them from scratch.
Sample-based single-path training is a common method for training the supernet, which is trained by uniform sampling or fair multipath sampling and optimizing single paths.
After training, the supernet can act as a performance estimator for different paths.
When choosing a path, it can be carried out through various search strategies, such as evolutionary algorithms or reinforcement learning.

\textbf{Predictor-based NAS Evaluators.}
The most popular predictor in NAS is the surrogate-based predictor \cite{sun2019surrogate} based on supervised learning. To obtain the data for training the predictor, it is necessary to train a large number of architectures initially, which is prohibitively time-consuming. The predictor then takes the architecture descriptions as inputs, and outputs the predicted performance scores. Utilizing a good predictor, promising architectures can be selected to be evaluated by the expensive evaluator. The query time is short, which allows large amount of predictions to be made during NAS.

In summary, both one-shot evaluators and predictor-based evaluators can accelerate the NAS process. However, how to combine them effectively and further reduce the computational time remains a challenging topic \cite{liu2022survey}.

\subsection{Evolutionary multiobjective Neural Architecture Search}
Aiming at automatically search for architectures with good performance in given search spaces, neural architecture search (NAS) methods make it possible to discover architectures without using a great deal of domain knowledge in comparison with designing networks manually.
Most NAS methods concentrate on discovering architectures with the best classification accuracy. However, in practical applications, other factors, such as model size, power consumption and robustness, should be taken into consideration at the same time.

Differentiable NAS (FBNet \cite{wu2019fbnet}), reinforcement learning based methods (MnasNet \cite{tan2019mnasnet}, MONAS \cite{hsu2018monas}) and Bayesian optimization (ChamNet \cite{dai2019chamnet}) usually transform multiobjective NAS into a single-objective one using scalarization or an additional constraint.
However, scalarized approaches were shown to be not as efficient as Pareto approaches. Multiobjective EAs are popular in solving multiobjective problems and have been shown to be successful in finding a set of Pareto optimal neural architectures in NAS.
To balance the accuracy and the inference speed, Kim \textit{et al}. proposed neuro-evolution with multiobjective optimization, dubbed NEMO, which employs the elitist non-dominated sorting genetic algorithm (NSGA-II) framework to search architectures.
Elsken \textit{et al}. \cite{elsken2018efficient} introduced LEMONADE that consider the validation error as expensive objective and model size as cheap objective. They evaluated cheap objective more times than expensive objective to save time. However, the experiment still requires 56 GPU days on CIFAR-10 datatset which is at a relatively high cost.
Yang \textit{et al}. \cite{yang2020cars} proposed pNSGA-III that simultaneously optimizes the model size and the classification accuracy.
Zhu \textit{et al}. \cite{zhu2019multi, zhu2021real} employed NSGA-II to optimize accuracy and floating-point operations (FLOPs) for federated NAS framework.
Hu \textit{et al}. \cite{hu2021accelerating} proposed random-weight evaluation to reduce the computational cost and adopted NSGA-II to optimize FLOPs and accuracy.

Several work have also considered two more objectives.
Lu \textit{et al}. \cite{lu2020nsganetv2} proposed NSGANet-V2 and they optimized five objectives simultaneously including accuracy, model size, MAdds, CPU and GPU latency.
Ning \textit{et al}. \cite{ning2020discovering} adopted a tournament-based EA framework and took performance and capacity into consideration, where the performance is measured by weighted sum of clean and adversarial accuracy. Other evolutionary multiobjective NAS for optimizing accuracy and robustness have introduced in Section \ref{secI}.

Above all, we can see that NSGA-II has been successfully employed in multiobjective NAS. Despite remarkable progress, the research on multiobjective NAS for the resilience of architectures against adversarial attacks has been sporadic. In our previous work, we introduced MORAS \cite{jia2021multi}, to search for architectures that are less sensitive to various adversarial attacks. However, the search process is time-consuming because each architecture in the population must be trained to obtain the fitness values. This work, by contrast, combines supernet training with surrogates to assist the evaluation process, thereby further improving the search efficiency.

\section{The Proposed approach} \label{iii}
\begin{figure*}[t]
	\centering
	\includegraphics[width=.98\textwidth]{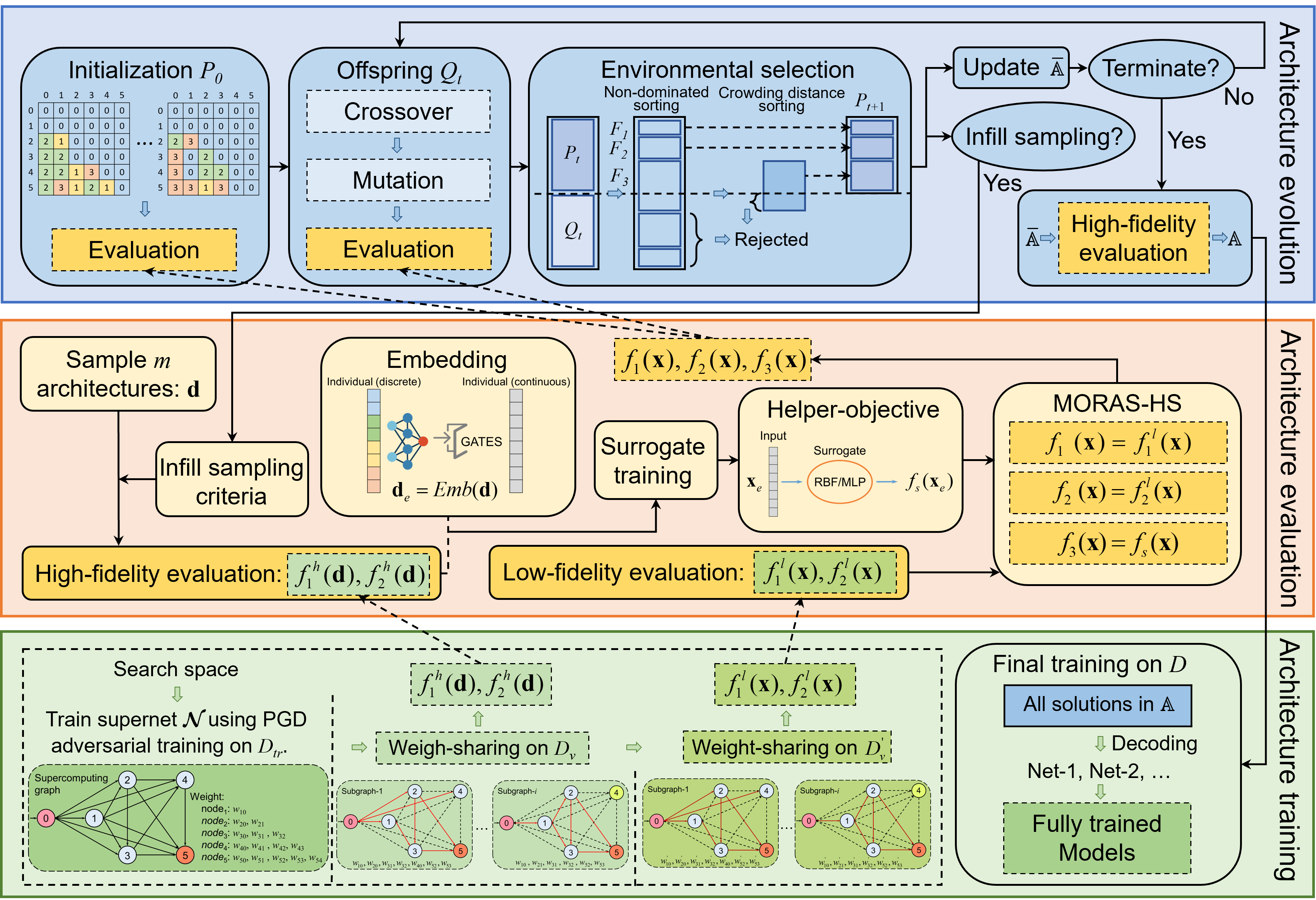}
	\caption{Overall Framework.
		Top: architecture evolution. This method employs NSGA-II as the main search framework. 
	Middle: architecture evaluation. This part is to efficiently evaluate the individuals obtained in the architecture evolution process, and the architecture relies on the supernet training and parameter sharing in the architecture training part. 
	Bottom: architecture training. This part includes supernet training, weight sharing, and a complete final training for the individuals resulting from the evolutionary process. 
	}
	\label{frm}
\end{figure*}
In this section, we develop a multiobjective architecture search for adversarial robustness with a surrogate as an auxiliary-objective, namely MORAS-SH. As shown in Figure \ref{frm}, MORAS-SH consists of three parts, i.e., architecture evolution, architecture evaluation, and architecture training.

We employ the elitist non-dominated sorting genetic algorithm (NSGA-II) \cite{deb2002fast} as the baseline for architecture evolution.
To efficiently evaluate the architectures, we estimate the performance of architectures on both clean images and adversarial examples by leveraging low-fidelity fitness evaluations. To guide the search for good solutions and help maintain diversity in the population, we train a surrogate by leveraging high-fidelity evaluations to predict the performance of the candidate architectures and the predicted value is used as a auxiliary-objective.
The low-fidelity and high-fidelity fitness values are obtained by inheriting weights $W$ from a pre-trained supernet $\mathcal{N}$ on partial and full validation sets, respectively.
\subsection{Preliminaries}

In this part, we will elaborate on the search space and encoding schemes in Section \ref{ss}. To make the prediction results of the surrogate model more accurate, we use GATES \cite{ning2020a} embedding method to convert the discrete variables into continuous ones, which will be introduced in Section \ref{emb}.

\begin{figure*}[htbp]
	\centering
	\includegraphics[width=.98\textwidth]{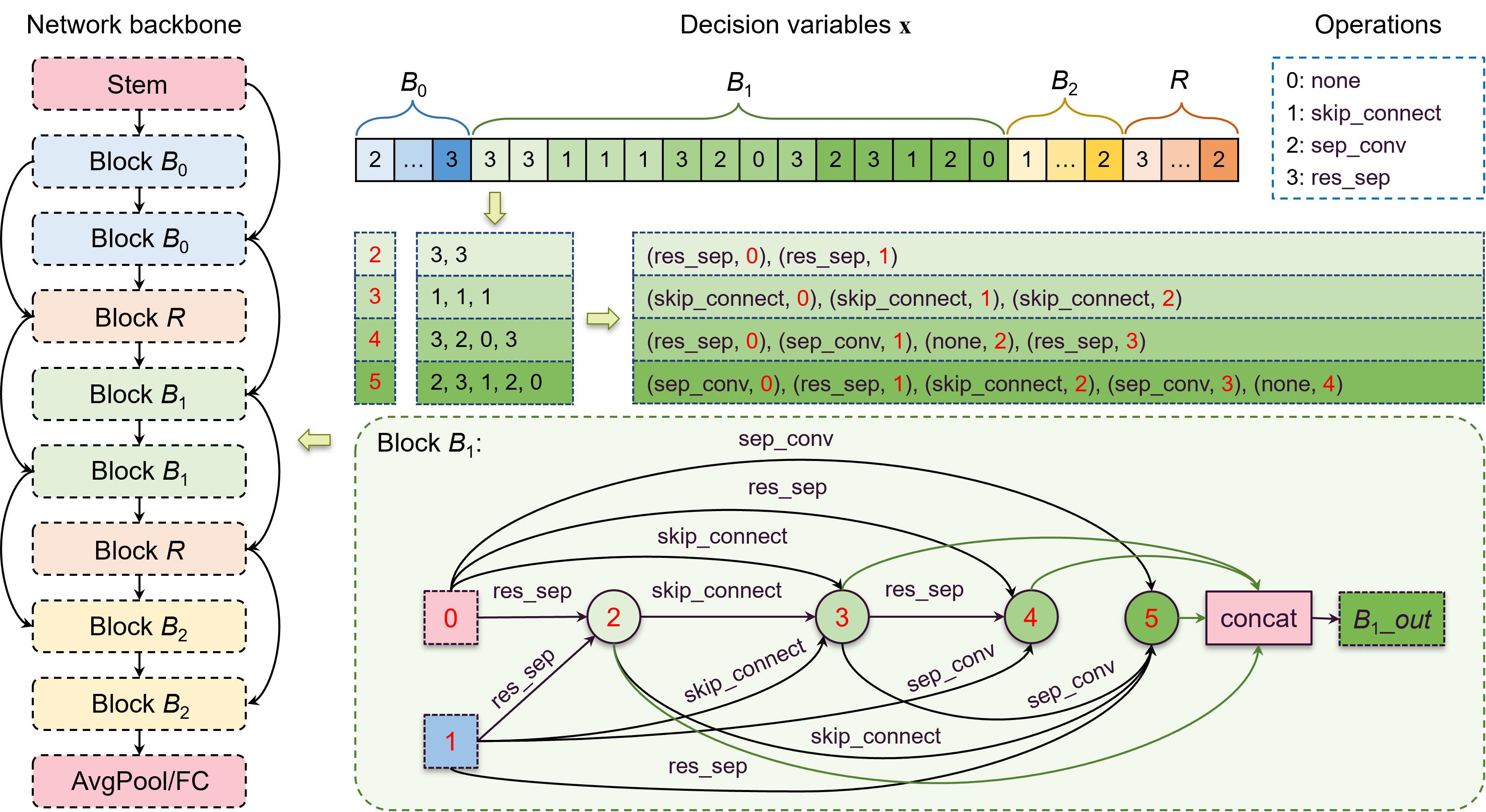}
	\caption{Network layout and encoding strategy.
	Left: overall cell-based architecture. Right: Decision variables, operations, the process of decoding Block $B_1$, and corresponding architecture of Block $B_1$.}
	\label{enc}
\end{figure*}
\subsubsection{Search Space and Encoding Strategy} \label{ss}

Expressive and appropriate search space is important for searching robust networks, since the type and size of the search space limits the range of possible networks that can be searched. The search space we use is the same as the stage-wise search space in \cite{ning2020discovering}. We use adjacency matrix encoding, which is the most common type of encodings used in current NAS research. Fig. \ref{enc} depicts the search space and encoding strategy. We search for four different blocks ($B_0$, $B_1$, $B_2$, and $R$). , the layout and the connections between each block are shown in the left part of Fig. \ref{enc}. The step size of Block $B_0$, $B_1$ and $B_2$ is one, and the step size of Block $R$ is two, that is, $R$ represents the reduction block.

Each block is represented as a directed acyclic graph (DAG) consisting of two nodes from the pre-previous block and the previous block, and four internal nodes in the current block.
Each internal node can receive information from the nodes of which the index number is less than its index (for example, node 3 can receive information from nodes 0, 1, 2), the edges between nodes represent operations, and the four optional operations are: none, skip connection, 3$\times$3 separable convolution, residual 3$\times$3 separable convolution. An example of the encoding scheme is given in Fig. \ref{enc}. The total dimension of the decision variable is $(2+3+4+5) \times 4=56$. Taking Block $B_1$ as an example, the first two dimensions of $B_1$ indicate that node 2 connects node 0 with operation 3 (res\_sep), and connects node 1 with operation 3 (res\_sep). The 3rd, 4th and 5th dimensions of $B_1$ indicate that nodes 3 connect node 0, node 1, and node 2 with operation 1 (skip\_connect), 1 (skip\_connect) and 1 (skip\_connect), respectively, and so on. The final output of Block $B_1$ (denoted as $B_1\_{out}$ in Fig. \ref{enc}) is obtained by concatenating all the internal nodes (we represent them with green lines).

\subsubsection{Embedding} \label{emb}
A surrogate model $s$, usually constructed by an MLP or RBF, takes a neural architecture as input and outputs a predicted score. However, recent studies show that if the coding representing the network architecture is directly used as the input of the predictor, the predicted results of the trained surrogate model are not accurate. Following \cite{ning2020discovering}, this work takes a graph-based neural architecture encoder called GATES \cite{ning2020a} that maps a neural architecture into a continuous embedding space, and then concatenate the embeddings of the four stage-wise block topologies ($B_0$, $B_1$, $B_2$, $R$) as the architecture embedding. The encoding process of GATES mimics the actual feature map computation, which can represent the information carried by the network.

Upon each operation $o$, GATES processed the input information $\mathbf{x}_{\text {in }}$ by a linear transform $W_{\mathbf{x}}$ and then element-wise multiplied with a soft attention mask.
$$
\mathbf{x}_{\mathrm{out}}=\mathbf{m} \odot \mathbf{x}_{\mathrm{in}} W_{\mathbf{x}}
$$
where $\odot$ denotes the element-wise multiplication, $W_{\mathbf{x}}$ is the transformation matrix on the information, and $\mathbf{m}$ is mask calculated from $\mathbf{m}=\sigma\left(\operatorname{EMB}(o) W_{o}\right) \in \mathbb{R}^{1 \times h_{i}}$, $\sigma(\cdot)$ is the sigmoid function, $EMB$ denotes the operation embedding, and $W_{o} \in \mathbb{R}^{h_{o} \times h_{i}}$ is a transformation matrix that transforms the $h_{o}$-dim operation embedding into a $h_{i}$-dim feature. The operation embedding $\operatorname{EMB}(o)=\operatorname{onehot}(o)^{T} \operatorname{EMB} \in \mathbb{R}^{1 \times h_{o}}$. Multiple pieces of information are aggregated at each node using summation. Finally, after obtaining the virtual information at all the nodes, the information at the output node is used as the embedding of the entire cell architecture.

\begin{figure*}[htb]
	\centering
	\includegraphics[width=.98\textwidth]{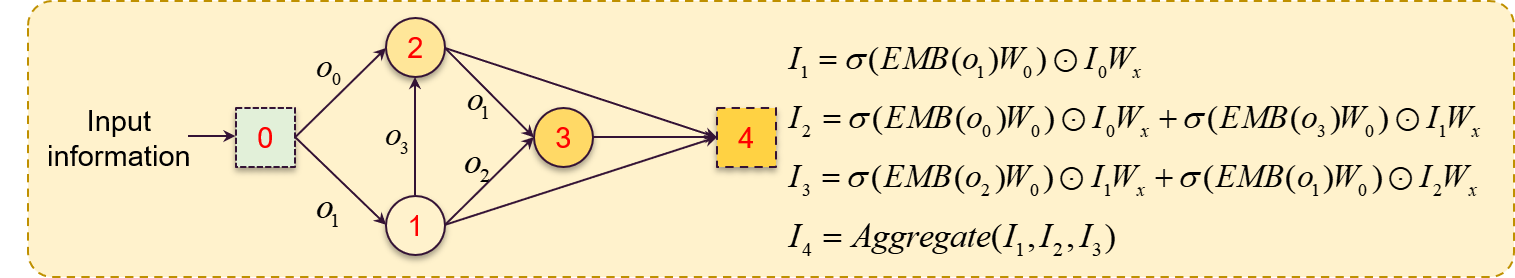}
	\caption{An example of GATES embedding. 
	Left: Architecture with node 0 as input, node 1, 2, and 3 as intermediate nodes, and node 4 as output. The operations are on edges.
	Right: $I_i$ denotes the embedding of node $i$.}
	\label{gates}
\end{figure*}

A schematic diagram of GATES is presented in Fig. \ref{gates}.
For example, the actual feature map computation at node 3 is $F_{3}= o_2(F_{1}) + o_1(F_{2})$, where $F_{i}$ is the feature map at node $i$.
GATES calculates the information at node 3 by $I_{3}=\sigma\left(E M B\left(o_{2}\right) W_{0}\right) \odot I_{1} W_{\mathbf{x}}+\sigma\left(E M B\left(o_{1}\right) W_{0}\right) \odot I_{2} W_{\mathbf{x}}$ to model the information processing of this feature map computation.


\subsection{Multiobjectivization} \label{mul}

Conceptually, multiobjectivization intends to simultaneously optimize the primary objective(s) together with additional auxiliary-objective(s) \cite{jensen2004helper}.
Since the auxiliary-objective(s) may help better guide the search, multiobjectivization may lead to better performance than merely focusing on the primary objective(s).
In this work, in addition to the primary objectives of accuracy and robustness of networks, we introduce the predicted score of a surrogate model as an auxiliary-objective . 
To the best of our knowledge, it is the first attempt to employ multiobjectivization in NAS by utilizing the predicted values of a surrogate model as an auxiliary-objective.

Both the evaluations using the low-fidelity fitness functions and surrogate models are computationally cheap yet correlated with the high-fidelity fitness function. However, neither of them is accurate enough to find a satisfactory solution to a bi-fidelity optimization problem. Moreover, the estimated performance according to the low-fidelity evaluation may be inconsistent with the one predicted by a surrogate. Hence, we use the predicted scores obtained from a surrogate as an additional objective to assist the evolutionary process with low-fidelity evaluation. We formulate NAS as the following three-objective minimization problem:

\begin{eqnarray}
  {\min}: F(\mathbf{x})= \{ {f_1},{f_2},{f_3}\}
\end{eqnarray}
\begin{equation}
    \begin{aligned}
	{f_{1}(\mathbf{x})} &= f_1^l(\mathbf{x}) = 1-(\frac{1}{N} \sum \mathbb{I}(\hat y==y))
	    \end{aligned}
\end{equation}
\begin{equation}
    \begin{aligned}
	{f_{2}(\mathbf{x})} &= f_2^l(\mathbf{x}) = 1-(\frac{1}{N} \sum \mathbb{I}(\hat y_{adv}==y))
    \end{aligned}
\end{equation}
\begin{equation}
    \begin{aligned}
	{f_{3}(\mathbf{x})} &= f_{s}(\mathbf{x})
    \end{aligned}
\end{equation}
where $f_{1}(\mathbf{x})$, $f_{2}(\mathbf{x})$ are the primary objectives,  with $\lbrace f_{1}^{l}(\mathbf{x})$, $f_{2}^{l}(\mathbf{x}) \rbrace$ denoting low-fidelity fitness evaluations calculated by the error rate on the partial validation set, and $f_{3}(\mathbf{x})$ represents the auxiliary-objective which is the predicted score of the surrogate model.

The surrogate model is trained to approximate high-fidelity fitness using data $\mathcal{S}$. Initially, we sample $m$ solutions using the Latin hypercube sampling (LHS) \cite{stein1987large} and calculate their fitness using high-fidelity evaluation, which is calculated by the error rate on the entire validation set.
Similar to low-fidelity evaluations, high-fidelity evaluations also yield error rates on clean data sets and adversarial samples. However, to simplify the optimization problem and reduce the difficulty of training the surrogate model, we employ a weighted sum of clean and adversarial error rates, rather than both of them, as a label for the architecture. That is, the predicted score and auxiliary-objective are only one dimension instead of two. Other existing work \cite{ning2020discovering} also uses this strategy when dealing with the accuracy of clean and adversarial datasets. The weights are both set to 0.5 for the sake of simplicity.

The inputs of the surrogate model are the values of architectures after embedding by using GATES \cite{ning2020a}.
The approximation error of the surrogate model is inevitable, which might misguide the search of the proposed algorithm. Therefore, infilling samples from the current population will be added to $\mathcal{S}$ after evolving $ G $ generations.
As suggested in \cite{wang2020transfer}, we select promising and uncertain solutions as infilling samples.
The pseudo-code of surrogate model construction is summarized in Algorithm \ref{alg1}.

\begin{algorithm}[htbp]\footnotesize{
		\caption{Pseudo code of surrogate model construction} \algblock{Begin}{End}
		\label{alg1}
		\textbf{Input:} $P_0$: Initial population; $ T $: the maximal generation number; $ G $: the maximal number of generations before updating a surrogate.\\
		\textbf{Output:} The non-dominated solutions in $\hat{\mathbb{A}}$.
		\begin{algorithmic}[1]
			\State \textbf{Initialization:} Set $\hat{\mathbb{A}}=\phi $, sample $ m $ individuals $ d $ to build $\mathcal{S}$ using LHS;
			\State \textbf{Evaluation:} Calculate the high-fidelity fitness $ \lbrace f_{1}^{h}(\mathbf{d}), f_{2}^{h}(\mathbf{d}) \rbrace $ of $\mathcal{S}$;
			\State $t = 0$;
			\For {$i=1$ to $\frac{T}{G}$}
    			\State \textbf{Embedding:} Encode cell architectures $d$ into embedding vectors $\mathbf{d_e}$;
    			\State \textbf{Training:}
    			
    			Train $f_{RBF}$ using  $\lbrace \mathcal{S}$,  $ 0.5 \times f_{h}^{1}(\mathbf{d_e}) +0.5\times  f_{h}^{2}(\mathbf{d_e}) \rbrace$;
    			
    			\For{$g=1$ to $G$}
        			\State \textbf{Estimating:} Evaluate $ P_t $ using  $F(\mathbf{x})$;
        			\State \textbf{Evolutionary process:}
        			 Generate offspring $Q_t$ by SBX and PM;
        			 Select $ n $ individuals from $P_t \cup Q_t$ using non-dominated sorting and crowding distance sorting for the next generation;
        			 Update $\hat{\mathbb{A}}$ with the individuals in the global first non-dominated ranking;
        			\State $t = t + 1$;
    			\EndFor
    			\If {$i \neq \frac{T}{G}$}
        			\State \textbf{Selection} Select $ k $ individuals $ \mathbf{x_c} $ from $ P_t $ according to criteria;
        			\State \textbf{Evaluation:} Evaluate $ x_c $ using $ f_{1}^{h}(\mathbf{x_c}), f_{2}^{h}(\mathbf{x_c})$;
        			\State \textbf{Updating $\mathcal{S}$:} $\mathcal{S}\leftarrow \mathcal{S} \cup \mathbf{x_c}$;
    			\EndIf
			\EndFor
	\end{algorithmic}}
\end{algorithm}
\subsection{Overall Framework}
The MORAS-SH workflow consists of three steps:
\subsubsection{Supernet training}
In a predefined architecture search space (Sec. \ref{ss}), we adversarially train a supernet $\mathcal{N}$ by using a 7-step PGD attack (PGD-7) (Sec. \ref{pgd}) on training data $D_{tr}$.
\subsubsection{Architecture evolution}
We randomly initialize a population $P_0$ with $n$ individuals (candidate topologies). For each individual, the low-fidelity evaluation is used to estimate its fitness values and the predicted score obtained from a surrogate model is used as an auxiliary-objective (Sec. \ref{mul}). The individuals in the population are gradually updated according to NSGA-II during the architecture optimization step. Concretely, we employ simulated binary crossover (SBX) and polynomial mutation (PM) \cite{deb1995simulated} to generate offspring. This process repeats $ G $ iterations and then $ k $ individuals from the current population are selected according to the infill criterion. The surrogate model will be updated using $\mathcal{S}$.
\subsubsection{Final training}
Since we evaluate the candidate architectures during the search process by using low-fidelity evaluation with the surrogate as an auxiliary-objective, the evaluations are of low precision. We consider this process as a pre-screening criterion. After the computation budget is exhausted, we evaluate all the non-dominant solutions in $\hat{\mathbb{A}}$ from the pre-screening criterion on the complete validation set with high fidelity to conduct secondary screening and then filter out the non-dominated solutions $\mathbb{A}$ for final adversarial training from scratch on fully training data $D$ using PGD-AT.

\section{Experiments} \label{iv}
The goal of the proposed MORAS-HS is to efficiently find the optimal architectures which achieve promising classification accuracy on both clean examples and adversarial examples. To this end, a series of experiments are designed in this section to demonstrate the advantage of the proposed method compared to its peer competitors. The following are the details of the experiments.

\subsection{Datasets}

Three widely-studied datasets are involved in the experiments, CIFAR-10 \cite{krizhevsky2010cifar}, CIFAR-100 \cite{krizhevsky2009learning} and Street View House Numbers (SVHN) \cite{netzer2011reading}.
We conduct a robust architecture search on CIFAR-10 and evaluate the discovered architectures on the CIFAR-10, CIFAR-100, and SVHN datasets.
CIFAR-10 and CIFAR-100 are labeled datasets that contain 10 and 100 classes, respectively. Both consist of a total number of 60,000 $ 32 \times 32$ pixel images. Therein, 50,000 images form the training set and the remaining from the test set.
SVHN data set consists of 10 digits obtained from real-world house numbers in Google Street View images.
It is composed of 630,420, $ 32 \times 32$ RGB color images in total, of which 73,257 samples are used for training, 26,032 for testing, and 531,131 additional less difficult samples to use as extra training data.

\subsection{Peer Competitors} \label{s4.2}
To demonstrate the superiority of the proposed approach, various peer competitors are selected for comparison, which can be divided into three different types.
\begin{itemize}
    \item The first group of baselines includes MobileNet-V2 \cite{sandler2018mobilenetv2}, VGG-16 \cite{simonyan2014very}, and ResNet-18 \cite{he2016deep}, which are manually designed by human experts.
    \item The second group represents NAS-based approaches in a search space that is similar to ours, including RobNet-Free \cite{guo2020meets}, MSRobNet-1560 \cite{ning2020discovering} and MSRobNet-1560-P \cite{ning2020discovering}.
    \item The third group of competitors is conducted for the ablation study. The main components of MORAS-SH include high-fidelity evaluation, low-fidelity evaluation, and surrogate modeling, which are actually part of the pre-screening of robust architectures. We conduct experiments with each component, which are termed MORAS-H, MORAS-L, MORAS-S.
\end{itemize}

\subsection{Implementation Details}
We used NVIDIA Titan RTX GPUs and implemented the experiments using PyTorch.
Before the search process, we divide the original training dataset of CIFAR-10 into two parts: training split (40000 images) and validation split (10000 images). The supernet is trained on the training split, and architecture rewards are evaluated on the validation split. PGD-7 under ${\ell}_{\infty}$ norm with $\epsilon=8/255$ and step size $\eta=2/255$ is used for adversarial training. We train the supernet with an initial channel number of 44 for 400 epochs. We use an SGD optimizer with a batch size of 64, a weight decay of 1e-4, and a momentum of 0.9. The learning rate is initially set to 0.05 and decayed to 0 following a cosine schedule.

We employ RBF and MLP as the surrogate during the search process separately. After GATES embedding, a 128-dimensional vector is fed into the surrogate. We sample 200 architectures to train an initial surrogate. The RBF model is built based on 128 Gaussian radial basis functions. To calculate the hyper-parameters of the RBF model using data $\mathcal{S}$ with $ m $ data points, the k-means clustering algorithm is adopted to obtain those 128 centers of radial basis functions, the maximum distances between the centers are set as the widths, and the pseudo-inverse method is used to assign the weights of Gaussian radial basis functions. We use a 3-layer MLP with 256 hidden units. In each surrogate training process, the MLP is trained for 100 epochs with batch size 50 and an Adam optimizer with a learning rate of 1e-3.

To limit the computational overhead, we set the maximum searching time as three days, which is very different from most existing work that uses the number of individuals being evaluated or limits the number of generations. The reason for this is that, first of all, it is difficult to have an exact value for the low-fidelity and high-fidelity evaluation time of the candidate architectures. If it is purely based on generation as a standard, it is more difficult to set parameters that can be fairly compared when doing other comparative experiments. In addition, the search time of the algorithm we mainly compared \cite{ning2020discovering} is about 1 GPU day, but its training time for the supernets is eight times that of ours (it takes about two days to train a super-network). Therefore, it is reasonable for us to set the termination condition to three days under this premise.

The portion of low-fidelity evaluation data is set to 0.2 according to \cite{zhou2021two}. To further alleviate the computational burden, we use the FGSM attack as an efficiency proxy \cite{ning2020discovering} of the PGD-7 since NAS does not necessarily require accurate performance, and the evaluation could be accelerated by roughly $8\times$.
We use a population size of 100 and update the surrogate every 20 iterations. Ten samples will be infilled to the set $\mathcal{S}$. The probabilities for crossover and mutation are set to 0.9 and 0.02, respectively.

For better performance, we augmented the initial channels of the architectures for the final training to 55.
For the final comparison on CIFAR-10, CIFAR-100 and SVHN, we adversarially train the architectures for 110 epochs on CIFAR-10/CIFAR-100 and 50 epochs on SVHN, using PGD-7 attacks with $\epsilon=8 / 255$ and step size $\eta=2 / 255$, and other settings are also kept the same.

To evaluate the adversarial robustness of the trained models, we apply the FGSM \cite{goodfellow2014explaining} with $\epsilon=8 / 255$, and PGD \cite{madry2018towards} with different step numbers.

\section{Experimental Results} \label{v}
We present the experimental results on CIFAR-10 in Section \ref{s5.1:c10} and the transferability on CIFAR-100 and SVHN datasets in Section \ref{s5.2}. The performance of the architectures obtained by the proposed MORAS-SH is compared to the first and second group of peer competitors mentioned in Section \ref{s4.2}. The ablation studies are presented in Section \ref{s5.3}, followed by discussion in Section \ref{s5.4}.

\subsection{Performance of MORAS-SH on CIFAR-10} \label{s5.1:c10}
\begin{table*}[htbp]
\caption{Comparison with peer competitors under various adversarial attacks on CIFAR-10.}
\label{res}
\centering
\resizebox{\textwidth}{!}{
\begin{tabular}{ccccccccc}
\hline
                                   & Architecture    & Clean (\%)    & FGSM (\%)     & PGD-7 (\%)    & PGD-20 (\%)   & PGD-100 (\%)  & \#Para (M) & FLOPS (M) \\ \hline
Manually designed networks         & MobileNet-V2    & 77.0          & 53.0          & 50.1          & 48.0          & 47.8          & 2.30       & 182       \\
                                   & VGG-16          & 79.9          & 53.7          & 50.4          & 48.1          & 47.9          & 14.73      & 626       \\
\textbf{}                          & ResNet-18       & 83.9          & 57.9          & 54.5          & 51.9          & 51.5          & 11.17      & 1110      \\ \hline
\multirow{3}{*}{NAS-based methods} & RobNet-Free     & 82.8          & 58.4          & 55.1          & 52.7          & 52.6          & 5.49       & 1560      \\
                                   & MSRobNet-1560   & 84.8          & \textbf{60.0} & \textbf{56.2} & \textbf{53.4} & \textbf{52.9} & 5.30       & 1588      \\
                                   & MSRobNet-1560-P & \textbf{85.2} & 59.4          & 55.2          & 51.9          & 51.5          & 4.88       & 1565      \\ \hline
\multirow{6}{*}{Ours}              & MORAS-SHNet-M1  & 85.8          & 59.4          & 55.5          & 52.5          & 52.1          & 5.22       & 1634      \\
                                   & MORAS-SHNet-M2  & 85.4          & \textbf{60.1} & 55.8          & 52.9          & 52.4          & 5.05       & 1606      \\
                                   & MORAS-SHNet-M3  & 85.5          & 59.6          & 55.6          & 52.8          & 52.5          & 5.20       & 1661      \\
                                   & MORAS-SHNet-R1  & \textbf{86.0} & 59.9          & 55.4          & 52.1          & 51.6          & 5.60       & 1525      \\
                                   & MORAS-SHNet-R2  & 85.6          & 59.9          & \textbf{56.2} & \textbf{53.1} & 52.6          & 5.42       & 1471      \\
                                   & MORAS-SHNet-R3  & 85.1          & 59.9          & 55.8          & 53.0          & \textbf{52.7} & 5.41       & 1484      \\ \hline
\end{tabular}}
\end{table*}
From the set of non-dominated solutions returned after the evolution, we obtained 91 and 43 architectures by using the proposed MORAS-SH with a surrogate of RBF and MLP, respectively. After the second screening by high-fidelity evaluation, the number of non-dominated solutions reduced to eight. We then fully train the 16 architectures and choose three architectures for each method based on their trade-offs.

Table \ref{res} compares the performances of the architectures under various adversarial attacks. The architectures discovered by our method are referred to as MORAS-SHNets, where MORAS-SHNet-M and MORAS-SHNet-R represent the architectures obtained by MORAS-SH with an MLP and RBF as the surrogate, respectively. The results of RobNets-Free and MORobNet series are extracted from \cite{guo2020meets} and \cite{ning2020discovering}, respectively.
The third column in Table \ref{res} represents the accuracy of each network on a clean test set. The fourth column indicates the accuracy of the networks on FGSM attacks. Columns 5, 6, and 7 indicate the accuracy of the networks on PGD-7, PGD-20, and PGD-100 attacks, respectively. The last two columns represent the number of parameters and FLOPS for the network, respectively. We indicate the best results of the competitors and our approach for each case in bold.

As seen from Table \ref{res}, our MORAS-SHNet-R1 achieves 86\% accuracy on clean data sets, outpacing all competitors. Under the FGSM attack, our MORAS-SHNet-M2 achieves an accuracy of 60.1\%, which is also the best result among the competitors. Under the PGD-7 attack, MORAS-SHNet-R2 achieves 56.2\% accuracy, the same as the MSRobNet-1560, which is also the highest. Under the PGD-20 and PGD-100 attacks, the best results are achieved by MSRobNet-1560, and our MORAS-SHNet-R2 and MORAS-SHNet-R3 are the second best.

We can see that the architectures discovered by MORAS-SHNets significantly outperform the manually designed CNNs under FGSM and strong adversarial attacks (PGD-7/10/100). With a similar number of parameters, MORAS-SHNets outperform other NAS-based peer competitors on clean samples and the samples with FGSM and PGD-7 attacks. Moreover, the computational cost of MORAS-SH is smaller than the MSRobNet series since only one supernet is trained instead of eight.
It illustrates that our approach can effectively and efficiently search for architectures with adversarial robustness.

\subsection{Transferability to CIFAR-100 and SVHN} \label{s5.2}
In line with the practice adopted in most previous NAS methods \cite{ning2020discovering}, we evaluate the transferability of the obtained architectures by inheriting the topology optimized for one dataset with weights retrained for a new dataset.
We train MORAS-SHNets on CIFAR-100 and SVHN, and show comparison in Table \ref{res-cifar100} and Table \ref{res-svhn}, respectively.

\begin{table*}[htbp]
\caption{Comparison with peer competitors under various adversarial attacks on CIFAR-100.}
\label{res-cifar100}
\centering
\resizebox{.8\textwidth}{!}{
\begin{tabular}{ccccccccc}
\hline
\multicolumn{1}{l}{\textbf{}}               & Architecture   & Clean (\%)    & FGSM (\%)     & PGD-7 (\%)    & PGD-20 (\%)   & PGD-100 (\%)  \\ \hline
\multirow{3}{*}{Manually designed networks} & MobileNet-V2   & 48.2          & 28.1          & 27.3          & 26.3          & 26.2          \\
                                            & VGG-16         & 51.5          & 29.1          & 27.1          & 25.8          & 25.8          \\
                                            & ResNet-18      & 59.2          & 33.8          & 31.6          & 29.9          & 29.7          \\ \hline
\multirow{3}{*}{NAS-based methods}          & RobNet-Free    & -             & -             & -             & -             & 23.9          \\
                                            & MSRobNet-1560  & 60.8          & \textbf{35.1} & \textbf{33.2} & \textbf{31.7} & \textbf{31.5} \\
                                            & MSRobNet-2000  & \textbf{61.6} & 34.8          & 32.9          & 31.6          & \textbf{31.5} \\ \hline
\multirow{6}{*}{Ours}                       & MORAS-SHNet-M1 & 61.4          & 32.9          & 30.5          & 28.6          & 28.4          \\
                                            & MORAS-SHNet-M2 & 61.2          & \textbf{34.1} & 30.9          & 29.1          & 28.8          \\
                                            & MORAS-SHNet-M3 & 61.5          & 33.9          & 32.6          & 29.5          & 29.3          \\
                                            & MORAS-SHNet-R1 & \textbf{61.8} & \textbf{34.1} & 30.8          & 28.6          & 28.2          \\
                                            & MORAS-SHNet-R2 & 61.4          & 33.0          & 30.6          & 28.9          & 28.5          \\
                                            & MORAS-SHNet-R3 & 61.4          & 33.0          & \textbf{33.1} & \textbf{31.3} & \textbf{31.2} \\ \hline
\end{tabular}}
\end{table*}

\begin{table*}[htbp]
\caption{Comparison with peer competitors under various adversarial attacks on SVHN.}
\label{res-svhn}
\centering
\resizebox{.8\textwidth}{!}{
\begin{tabular}{ccccccccc}
\hline
\multicolumn{1}{l}{}                        & Architecture   & Clean (\%)    & FGSM (\%)     & PGD-7 (\%)    & PGD-20 (\%)   & PGD-100 (\%)  \\ \hline
\multirow{3}{*}{Manually designed networks} & MobileNet-V2   & 93.9          & 73.0          & 61.9          & 55.7          & 53.9          \\
                                            & VGG-16         & 92.3          & 66.6          & 55.0          & 47.4          & 45.1          \\
                                            & ResNet-18      & 92.3          & 73.5          & 57.4          & 51.2          & 48.8          \\ \hline
\multirow{3}{*}{NAS-based methods}          & RobNet-Free    & 94.2          & 84.0          & \textbf{66.1} & \textbf{59.7} & \textbf{56.9} \\
                                            & MSRobNet-1560  & \textbf{95.0} & 77.5          & 64.0          & 57.0          & 54.2          \\
                                            & MSRobNet-2000  & 94.9          & \textbf{84.8} & 65.3          & 58.8          & 55.1          \\ \hline
\multirow{6}{*}{Ours}                       & MORAS-SHNet-M1 & 94.8          & 86.7          & 78.4          & 66.0          & 61.2          \\
                                            & MORAS-SHNet-M2 & 94.4          & 84.3          & 65.3          & 58.6          & 55.6          \\
                                            & MORAS-SHNet-M3 & \textbf{95.8} & \textbf{90.6} & \textbf{85.7} & \textbf{73.7} & \textbf{66.3} \\
                                            & MORAS-SHNet-R1 & 94.9          & 85.4          & 64.1          & 57.8          & 54.9          \\
                                            & MORAS-SHNet-R2 & 94.3          & 83.9          & 63.8          & 58.1          & 55.4          \\
                                            & MORAS-SHNet-R3 & 94.7          & 77.3          & 61.4          & 55.1          & 52.8          \\ \hline
\end{tabular}}
\end{table*}
In general, our models are consistently more robust than manually designed networks on both CIFAR-100 and SVHN. As shown in Table \ref{res-cifar100}, our MORAS-SHNet-R1 outperforms others on clean CIFAR-100. However, under different adversarial attacks, MSRobNet-1560 achieves the best results.

As reported in Table \ref{res-svhn}, our MORAS-SHNet-M3 outperforms all peer competitors in all cases on SVHN. The results under different adversarial attacks are much better than the peer competitors.
\subsection{Ablation Study} \label{s5.3}

This section aims to disentangle the individual contribution of each principal component in the proposed method. In the ablation experiment, we assume that the fitness values obtained by the parameters of the architectures inherited directly from the supernet after a complete validation set test are relatively accurate. That is, we use high-fidelity evaluation to measure the performance of the algorithms under comparison in the pre-screening process. To be fair, all experiments are terminated for three days.

\begin{figure}[t]
	\centering
	\includegraphics[width=.48\textwidth]{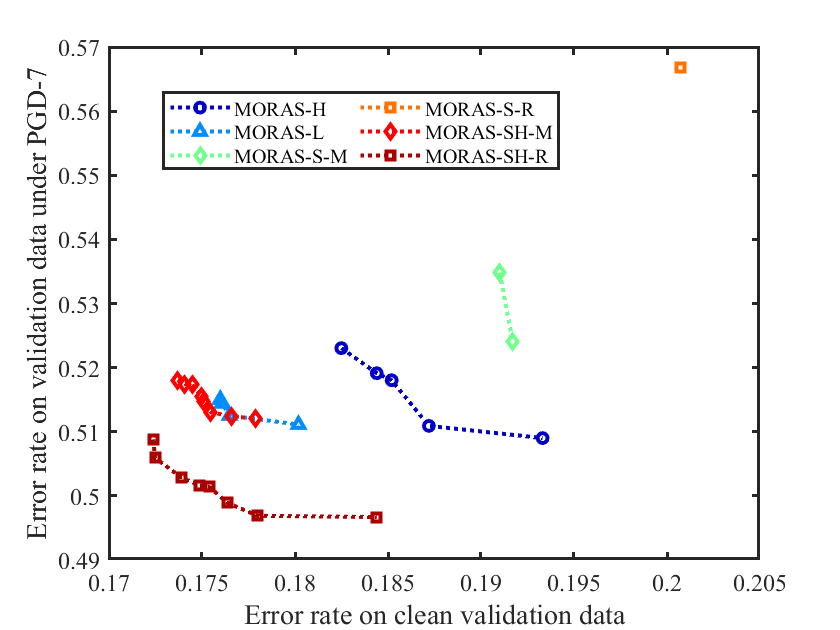}
	\caption{Pareto fronts obtained by comparative experiments. The parameters are inherited from the supernet.}
	\label{pf}
\end{figure}

\begin{table}[htbp]
\caption{The number of non-dominated solutions on pre-screening and secondary screening.}
\label{num_nd}
\centering
\resizebox{.48\textwidth}{!}{
\begin{tabular}{ccc}
\hline
           & Pre-screening & Secondary screening \\ \hline
MORAS-H    & 5             & 5                   \\
MORAS-L    & 13            & 4                   \\
MORAS-S-M  & 18            & 2                   \\
MORAS-S-R  & 4             & 3                   \\
MORAS-SH-M & 43            & 8                   \\
MORAS-SH-R & 91            & 8                   \\ \hline
\end{tabular}}
\end{table}
Over a three-day evolutionary process, we obtained the non-dominated solutions of each experiment for pre-screening, the number of which we showed in the first column of Table \ref{num_nd}. As we can see from Table \ref{num_nd}, the number of solutions obtained solely using high-fidelity, low-fidelity, or surrogate models is small. The method we propose, with the surrogate model as an auxiliary-objective, has a large number of solutions obtained in the pre-screening. In the secondary screening, the number of non-dominated solutions for each experiment is listed in the second column of Table \ref{num_nd} after further high-fidelity evaluation. Due to the different generations and evaluation methods of the various comparison experiments, we could not compare their HV curves. Here, we draw the Pareto frontier obtained after a high-fidelity evaluation of the predicted non-dominated solution obtained after three days of running each experiment on Fig. \ref{pf}.

\begin{figure}[t]
	\centering
	\includegraphics[width=.48\textwidth]{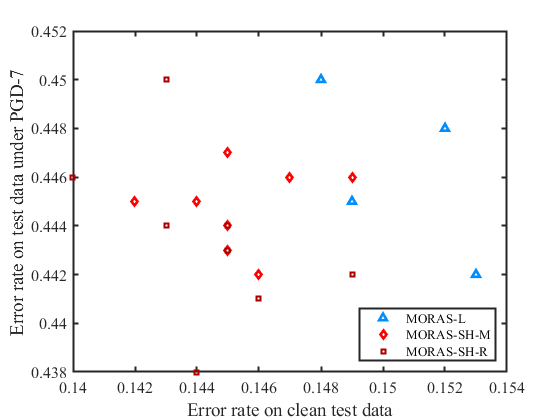}
	\caption{The performance of architectures obtained by comparative experiments after adversarial training from scratch.}
	\label{nd-fn}
\end{figure}

As can be seen from Fig. \ref{pf}, the solutions obtained by MORAS-H, MORAS-L, MORAS-S-M, and MORAS-S-R are dominated by the solutions obtained by the methods we propose. The solutions obtained by MORAS-L are comparable to our method. In order to prove the superiority of our method, we further trained them from scratch in a complete adversarial training on CIFAR-10. We show the performance of each network after training in Figure \ref{nd-fn}. As shown in Figure \ref{nd-fn}, the architectures that MORAS-L searched for are dominated by most of the architectures we searched. To further compare the performance of the network obtained by each method, we also listed the results of the network on various attacks, as shown in Table \ref{tbl5}. As can be seen from Table \ref{tbl5}, our approach is still superior to MORAS-L, whether it is FGSM or under stronger attacks. This further validates the effectiveness of our surrogate model as an auxiliary-objective.

Here we explain why separately considering different parts fails to obtain promising Pareto fronts.
It is known that evaluating each candidate architecture through high-fidelity evaluations at each iteration is prohibitively expensive. With a given budget, MORAS-H can iterate only a few generations, leading to MORAS-H's inability to find better architectures in a vast search space.
If using low-fidelity evaluation merely during the search process, MORAS-L can search for more generations in a limited time budget. However, the diversity of solutions is poor, and the number of non-dominated solutions obtained after the search is also small.
If using the surrogate model merely in the search process, the search will be misled due to the inaccurate prediction results of the previous surrogate model. Even if the surrogate model is gradually updated with the generation increases, the range of predicted values of the updated surrogate model also changes. This means that if the surrogate model becomes more accurate, but its predictions are not as good as the previous generation of individuals, such excellent individuals will also be eliminated. In other work \cite{ning2020a,ning2020discovering}, researchers have trained a surrogate model with the relative ranking rather than the absolute performance values, which does not arise from the above problems. In future studies, we will introduce relative ranking into multiobjective evolutionary searches.
\begin{table*}[htbp]
\caption{Comparison with MORAS-L-Nets under various adversarial attacks on CIFAR-10.}
\label{tbl5}
\centering
\resizebox{.8\textwidth}{!}{
\begin{tabular}{cccccccc}
\hline
Architecture   & Clean (\%)    & FGSM (\%)     & PGD-7 (\%)    & PGD-20 (\%)   & PGD-100 (\%)  & \#Para (M) & FLOPS (M) \\ \hline
MORAS-L-Net1  & 84.7          & 59.1          & 55.8          & 52.7          & 52.3          & 4.82       & 1426      \\
MORAS-L-Net2  & 84.8          & 58.9          & 55.2          & 52.2          & 52.0          & 4.88       & 1453      \\
MORAS-L-Net3  & 85.2          & 59.1          & 55.0          & 52.0          & 51.7          & 5.12       & 1481      \\
MORAS-L-Net4  & 85.1          & 59.6          & 55.5          & 52.6          & 52.1          & 5.09       & 1508      \\ \hline
MORAS-SHNet-M1 & 85.8          & 59.4          & 55.5          & 52.5          & 52.1          & 5.22       & 1634      \\
MORAS-SHNet-M2 & 85.4          & \textbf{60.1} & 55.8          & 52.9          & 52.4          & 5.05       & 1606      \\
MORAS-SHNet-M3 & 85.5          & 59.6          & 55.6          & 52.8          & 52.5          & 5.20       & 1661      \\
MORAS-SHNet-M4 & 85.1          & 59.3          & 55.4          & 52.4          & 52.0          & 5.27       & 1660      \\
MORAS-SHNet-M5 & 85.6          & 60.0          & 55.5          & 52.5          & 52.1          & 5.05       & 1606      \\
MORAS-SHNet-M6 & 85.3          & 59.1          & 55.4          & 52.3          & 51.8          & 5.22       & 1634      \\
MORAS-SHNet-M7 & 85.5          & 59.4          & 55.3          & 52.2          & 51.7          & 5.07       & 1635      \\
MORAS-SHNet-M8 & 85.5          & 60.0          & 55.7          & 52.8          & 52.3          & 5.14       & 1634      \\ \hline
MORAS-SHNet-R1 & \textbf{86.0} & 59.9          & 55.4          & 52.1          & 51.6          & 5.6        & 1525      \\
MORAS-SHNet-R2 & 85.6          & 59.9          & \textbf{56.2} & \textbf{53.1} & 52.6          & 5.42       & 1471      \\
MORAS-SHNet-R3 & 85.1          & 59.9          & 55.8          & 53.0          & \textbf{52.7} & 5.41       & 1484      \\
MORAS-SHNet-R4 & 85.7          & 59.9          & 55.6          & 52.3          & 51.9          & 4.85       & 1397      \\
MORAS-SHNet-R5 & 85.7          & 59.3          & 55.0          & 52.0          & 51.7          & 4.93       & 1397      \\
MORAS-SHNet-R6 & 85.5          & 59.6          & 55.6          & 52.4          & 51.9          & 4.99       & 1409      \\
MORAS-SHNet-R7 & 85.4          & 59.8          & 55.9          & 52.9          & 52.6          & 5.06       & 1367      \\
MORAS-SHNet-R8 & 85.5          & 59.7          & 55.7          & 52.6          & 52.2          & 5.29       & 1445      \\ \hline
\end{tabular}}
\end{table*}
\subsection{Discussion} \label{s5.4}
We combined weight sharing and a surrogate-assisted approach to search for robust network architectures at a limited computational cost. Using the surrogate model as an additional objective, we found that our approach can efficiently and effectively search for robust architectures compared to peer competitors on the CIFAR-10 dataset. Moreover, the network architecture is also transferable, especially on the SVHN dataset.

In terms of computational cost, we only pre-train one supernet at a time and then inherit the parameters of the supernet when evaluating the candidate architecture, which significantly reduces the time required to evaluate the performance of the network. Moreover, we used a combination of low-fidelity evaluations and surrogate models to further speed up the search efficiency. The performance of the network obtained by our method is comparable to that of MSRobNets, but the computational cost is much less, since MSRobNets need to train eight supernets. It demonstrates that our approach can efficiently search for robust networks.

The time complexity of NSGA-II per generation is $O(MN^2)$. The number of objectives $M$ is two if we merely consider the primary objectives. The computational complexity will increase to $O(4N^2)$ if we consider accuracy and adversarial robustness predicted by the surrogate model separately. In this work, we employ a weighted sum of the clean and adversarial error rates as a label for the architecture to simplify the optimization problem and reduce the difficulty of training the surrogate model. That is, the predicted score and auxiliary-objective are only one dimension instead of two. The weights are both set to 0.5 for the sake of simplicity. Therefore, the time complexity is $O(3N^2)$.

Regarding transferability, our network also achieved better performance after training from scratch on CIFAR-100 and SVHN datasets. Especially on the SVHN dataset, the results are far superior to other networks. On the CIFAR-100, the overall performance of our network is not as good as MSRobNet. Therefore, the transferability of this method needs to be improved. In the future, we will also do some research to enhance the transferability of the network.

\section{Conclusion} \label{vi}

We employ an MOEA-based NAS approach to search for architectures that are robust to adversarial attacks.
To make the procedure efficient, we propose a multiobjective architecture search for adversarial robustness with the assistance of a surrogate model as an auxiliary-objective, namely, MORAS-SH.
During evolution, MORAS-SH maximally utilizes the learned knowledge from both low- and high-fidelity fitness.
Experiments results on benchmark datasets demonstrate that the proposed MORAS-SH can efficiently provide several architectures on the Pareto front. The searched models are also superior to peer competitors in terms of robustness and accuracy.

Most research on NAS for robust architectures focuses on network architectures that perform well on both clean and adversarial examples but ignore those that perform well on clean data sets but are sensitive to attacks. Few researchers have studied what kind of network topology or parameters cause this phenomenon. It is an exciting topic to study networks that perform well on clean data but are sensitive to attacks, and it will help to further understand the intrinsic nature of neural networks. Early detection of structural factors that make networks sensitive can accelerate the discovery and design of more robust networks.

\ifCLASSOPTIONcaptionsoff
  \newpage
\fi



\bibliographystyle{IEEEtran}
\bibliography{my_refs}

\begin{thebibliography}{10}
\providecommand{\url}[1]{#1}
\csname url@samestyle\endcsname
\providecommand{\newblock}{\relax}
\providecommand{\bibinfo}[2]{#2}
\providecommand{\BIBentrySTDinterwordspacing}{\spaceskip=0pt\relax}
\providecommand{\BIBentryALTinterwordstretchfactor}{4}
\providecommand{\BIBentryALTinterwordspacing}{\spaceskip=\fontdimen2\font plus
\BIBentryALTinterwordstretchfactor\fontdimen3\font minus
  \fontdimen4\font\relax}
\providecommand{\BIBforeignlanguage}[2]{{%
\expandafter\ifx\csname l@#1\endcsname\relax
\typeout{** WARNING: IEEEtran.bst: No hyphenation pattern has been}%
\typeout{** loaded for the language `#1'. Using the pattern for}%
\typeout{** the default language instead.}%
\else
\language=\csname l@#1\endcsname
\fi
#2}}
\providecommand{\BIBdecl}{\relax}
\BIBdecl

\bibitem{krizhevsky2012imagenet}
A.~Krizhevsky, I.~Sutskever, and G.~E. Hinton, ``Imagenet classification with
  deep convolutional neural networks,'' in \emph{Advances in Neural Information
  Processing Systems}, 2012, pp. 1097--1105.

\bibitem{zeiler2014visualizing}
M.~D. Zeiler and R.~Fergus, ``Visualizing and understanding convolutional
  networks,'' in \emph{European Conference on Computer Vision}.\hskip 1em plus
  0.5em minus 0.4em\relax Springer, 2014, pp. 818--833.

\bibitem{lin2013network}
M.~Lin, Q.~Chen, and S.~Yan, ``Network in network,'' \emph{arXiv preprint
  arXiv:1312.4400}, 2013.

\bibitem{simonyan2014very}
\BIBentryALTinterwordspacing
K.~Simonyan and A.~Zisserman, ``Very deep convolutional networks for
  large-scale image recognition,'' in \emph{3rd International Conference on
  Learning Representations, {ICLR} 2015, San Diego, CA, USA, May 7-9, 2015,
  Conference Track Proceedings}, Y.~Bengio and Y.~LeCun, Eds., 2015. [Online].
  Available: \url{http://arxiv.org/abs/1409.1556}
\BIBentrySTDinterwordspacing

\bibitem{szegedy2015going}
C.~Szegedy, W.~Liu, Y.~Jia, P.~Sermanet, S.~Reed, D.~Anguelov, D.~Erhan,
  V.~Vanhoucke, and A.~Rabinovich, ``Going deeper with convolutions,'' in
  \emph{Proceedings of the IEEE Conference on Computer Vision and Pattern
  Recognition}, 2015, pp. 1--9.

\bibitem{he2016deep}
K.~He, X.~Zhang, S.~Ren, and J.~Sun, ``Deep residual learning for image
  recognition,'' in \emph{Proceedings of the IEEE Conference on Computer Vision
  and Pattern Recognition}, 2016, pp. 770--778.

\bibitem{huang2017densely}
G.~Huang, Z.~Liu, L.~Van Der~Maaten, and K.~Q. Weinberger, ``Densely connected
  convolutional networks,'' in \emph{Proceedings of the IEEE Conference on
  Computer Vision and Pattern Recognition}, 2017, pp. 4700--4708.

\bibitem{sabour2017dynamic}
S.~Sabour, N.~Frosst, and G.~E. Hinton, ``Dynamic routing between capsules,''
  in \emph{Advances in Neural Information Processing Systems}, 2017, pp.
  3856--3866.

\bibitem{ren2015faster}
S.~Ren, K.~He, R.~Girshick, and J.~Sun, ``Faster {R-CNN}: Towards real-time
  object detection with region proposal networks,'' \emph{IEEE Transactions on
  Pattern Analysis and Machine Intelligence}, vol.~39, no.~6, pp. 1137--1149,
  2017.

\bibitem{ren2016object}
S.~Ren, K.~He, R.~Girshick, X.~Zhang, and J.~Sun, ``Object detection networks
  on convolutional feature maps,'' \emph{IEEE Transactions on Pattern Analysis
  and Machine Intelligence}, vol.~39, no.~7, pp. 1476--1481, 2017.

\bibitem{sutskever2014sequence}
I.~Sutskever, O.~Vinyals, and Q.~V. Le, ``Sequence to sequence learning with
  neural networks,'' in \emph{Advances in Neural Information Processing
  Systems}, Z.~Ghahramani, M.~Welling, C.~Cortes, N.~Lawrence, and
  K.~Weinberger, Eds., vol.~27.\hskip 1em plus 0.5em minus 0.4em\relax Curran
  Associates, Inc., 2014.

\bibitem{szegedy2013intriguing}
C.~Szegedy, W.~Zaremba, I.~Sutskever, J.~Bruna, D.~Erhan, I.~Goodfellow, and
  R.~Fergus, ``Intriguing properties of neural networks,'' \emph{arXiv preprint
  arXiv:1312.6199}, 2013.

\bibitem{yuan2019adversarial}
X.~Yuan, P.~He, Q.~Zhu, and X.~Li, ``Adversarial examples: Attacks and defenses
  for deep learning,'' \emph{IEEE Transactions on Neural Networks and Learning
  Systems}, vol.~30, no.~9, pp. 2805--2824, 2019.

\bibitem{he2016identity}
K.~He, X.~Zhang, S.~Ren, and J.~Sun, ``Identity mappings in deep residual
  networks,'' in \emph{European Conference on Computer Vision}.\hskip 1em plus
  0.5em minus 0.4em\relax Springer, 2016, pp. 630--645.

\bibitem{zagoruyko2016wide}
S.~Zagoruyko and N.~Komodakis, ``Wide residual networks,'' in \emph{British
  Machine Vision Conference 2016}.\hskip 1em plus 0.5em minus 0.4em\relax
  British Machine Vision Association, 2016.

\bibitem{zoph2016neural}
B.~Zoph and Q.~V. Le, ``Neural architecture search with reinforcement
  learning,'' \emph{arXiv preprint arXiv:1611.01578}, 2016.

\bibitem{pham2018efficient}
H.~Pham, M.~Guan, B.~Zoph, Q.~Le, and J.~Dean, ``Efficient neural architecture
  search via parameters sharing,'' in \emph{International Conference on Machine
  Learning}.\hskip 1em plus 0.5em minus 0.4em\relax PMLR, 2018, pp. 4095--4104.

\bibitem{real2019regularized}
E.~Real, A.~Aggarwal, Y.~Huang, and Q.~V. Le, ``Regularized evolution for image
  classifier architecture search,'' in \emph{Proceedings of the AAAI Conference
  on Artificial Intelligence}, vol.~33, no.~01, 2019, pp. 4780--4789.

\bibitem{wu2019fbnet}
B.~Wu, X.~Dai, P.~Zhang, Y.~Wang, F.~Sun, Y.~Wu, Y.~Tian, P.~Vajda, Y.~Jia, and
  K.~Keutzer, ``Fbnet: Hardware-aware efficient convnet design via
  differentiable neural architecture search,'' in \emph{Proceedings of the
  IEEE/CVF Conference on Computer Vision and Pattern Recognition}, 2019, pp.
  10\,734--10\,742.

\bibitem{cai2018efficient}
H.~Cai, T.~Chen, W.~Zhang, Y.~Yu, and J.~Wang, ``Efficient architecture search
  by network transformation,'' in \emph{Proceedings of the AAAI Conference on
  Artificial Intelligence}, vol.~32, no.~1, 2018.

\bibitem{jia2021multi}
J.~Liu and Y.~Jin, ``Multi-objective search of robust neural architectures
  against multiple types of adversarial attacks,'' \emph{Neurocomputing}, vol.
  453, pp. 73--84, 2021.

\bibitem{cai2019once}
\BIBentryALTinterwordspacing
H.~Cai, C.~Gan, T.~Wang, Z.~Zhang, and S.~Han, ``Once-for-all: Train one
  network and specialize it for efficient deployment,'' in \emph{International
  Conference on Learning Representations}, 2020. [Online]. Available:
  \url{https://openreview.net/forum?id=HylxE1HKwS}
\BIBentrySTDinterwordspacing

\bibitem{liu2018progressive}
C.~Liu, B.~Zoph, M.~Neumann, J.~Shlens, W.~Hua, L.-J. Li, L.~Fei-Fei,
  A.~Yuille, J.~Huang, and K.~Murphy, ``Progressive neural architecture
  search,'' in \emph{Proceedings of the European Conference on Computer Vision
  (ECCV)}, 2018, pp. 19--34.

\bibitem{sun2019surrogate}
Y.~Sun, H.~Wang, B.~Xue, Y.~Jin, G.~G. Yen, and M.~Zhang, ``Surrogate-assisted
  evolutionary deep learning using an end-to-end random forest-based
  performance predictor,'' \emph{IEEE Transactions on Evolutionary
  Computation}, vol.~24, no.~2, pp. 350--364, 2019.

\bibitem{chen2020multi}
Z.~Chen, F.~Zhou, G.~Trimponias, and Z.~Li, ``Multi-objective neural
  architecture search via non-stationary policy gradient,'' \emph{arXiv
  preprint arXiv:2001.08437}, 2020.

\bibitem{luo2018neural}
R.~Luo, F.~Tian, T.~Qin, E.~Chen, and T.-Y. Liu, ``Neural architecture
  optimization,'' in \emph{Proceedings of the 32nd International Conference on
  Neural Information Processing Systems}, 2018, pp. 7827--7838.

\bibitem{alparslan2021atras}
Y.~Alparslan and E.~Kim, ``{ATRAS}: Adversarially trained robust architecture
  search,'' \emph{arXiv preprint arXiv:2106.06917}, 2021.

\bibitem{huang2021exploring}
H.~Huang, Y.~Wang, S.~Erfani, Q.~Gu, J.~Bailey, and X.~Ma, ``Exploring
  architectural ingredients of adversarially robust deep neural networks,''
  \emph{Advances in Neural Information Processing Systems}, vol.~34, pp.
  5545--5559, 2021.

\bibitem{vargas2019evolving}
D.~V. Vargas and S.~Kotyan, ``Evolving robust neural architectures to defend
  from adversarial attacks,'' \emph{arXiv preprint arXiv:1906.11667}, 2019.

\bibitem{dong2019neural}
N.~Dong, M.~Xu, X.~Liang, Y.~Jiang, W.~Dai, and E.~Xing, ``Neural architecture
  search for adversarial medical image segmentation,'' in \emph{International
  Conference on Medical Image Computing and Computer-Assisted
  Intervention}.\hskip 1em plus 0.5em minus 0.4em\relax Springer, 2019, pp.
  828--836.

\bibitem{guo2020meets}
M.~Guo, Y.~Yang, R.~Xu, Z.~Liu, and D.~Lin, ``When {NAS} meets robustness: In
  search of robust architectures against adversarial attacks,'' in
  \emph{Proceedings of the IEEE/CVF Conference on Computer Vision and Pattern
  Recognition}, 2020, pp. 631--640.

\bibitem{yue2020effective}
Z.~Yue, B.~Lin, X.~Huang, and Y.~Zhang, ``Effective, efficient and robust
  neural architecture search,'' \emph{arXiv preprint arXiv:2011.09820}, 2020.

\bibitem{ning2020discovering}
X.~Ning, J.~Zhao, W.~Li, T.~Zhao, Y.~Zheng, H.~Yang, and Y.~Wang, ``Discovering
  robust convolutional architecture at targeted capacity: A multi-shot
  approach,'' \emph{arXiv preprint arXiv:2012.11835}, 2020.

\bibitem{devaguptapu2021adversarial}
C.~Devaguptapu, D.~Agarwal, G.~Mittal, P.~Gopalani, and V.~N. Balasubramanian,
  ``On adversarial robustness: A neural architecture search perspective,'' in
  \emph{Proceedings of the IEEE/CVF International Conference on Computer
  Vision}, 2021, pp. 152--161.

\bibitem{chen2020anti}
H.~Chen, B.~Zhang, S.~Xue, X.~Gong, H.~Liu, R.~Ji, and D.~Doermann,
  ``Anti-bandit neural architecture search for model defense,'' in
  \emph{European Conference on Computer Vision}.\hskip 1em plus 0.5em minus
  0.4em\relax Springer, 2020, pp. 70--85.

\bibitem{cazenavette2021architectural}
G.~Cazenavette, C.~Murdock, and S.~Lucey, ``Architectural adversarial
  robustness: The case for deep pursuit,'' in \emph{Proceedings of the IEEE/CVF
  Conference on Computer Vision and Pattern Recognition (CVPR)}, June 2021, pp.
  7150--7158.

\bibitem{Hosseini_2021_CVPR}
R.~Hosseini, X.~Yang, and P.~Xie, ``Dsrna: Differentiable search of robust
  neural architectures,'' in \emph{Proceedings of the IEEE/CVF Conference on
  Computer Vision and Pattern Recognition (CVPR)}, June 2021, pp. 6196--6205.

\bibitem{wang2021neural}
K.~Wang, P.~Xu, C.-M. Chen, S.~Kumari, M.~Shojafar, and M.~Alazab, ``Neural
  architecture search for robust networks in 6g-enabled massive iot domain,''
  \emph{IEEE Internet of Things Journal}, vol.~8, no.~7, pp. 5332--5339, 2021.

\bibitem{mok2021advrush}
J.~Mok, B.~Na, H.~Choe, and S.~Yoon, ``Advrush: Searching for adversarially
  robust neural architectures,'' in \emph{Proceedings of the IEEE/CVF
  International Conference on Computer Vision}, 2021, pp. 12\,322--12\,332.

\bibitem{liu2018darts}
H.~Liu, K.~Simonyan, and Y.~Yang, ``{DARTS}: Differentiable architecture
  search,'' in \emph{International Conference on Learning Representations},
  2019.

\bibitem{xie2021tiny}
G.~Xie, J.~Wang, G.~Yu, F.~Zheng, and Y.~Jin, ``Tiny adversarial
  mulit-objective oneshot neural architecture search,'' \emph{arXiv preprint
  arXiv:2103.00363}, 2021.

\bibitem{goodfellow2014explaining}
I.~J. Goodfellow, J.~Shlens, and C.~Szegedy, ``Explaining and harnessing
  adversarial examples,'' in \emph{International Conference on Learning
  Representations}, 2015.

\bibitem{kurakin2016adversarial}
H.~Ren and T.~Huang, ``Adversarial example attacks in the physical world,'' in
  \emph{Machine Learning for Cyber Security}, X.~Chen, H.~Yan, Q.~Yan, and
  X.~Zhang, Eds.\hskip 1em plus 0.5em minus 0.4em\relax Cham: Springer
  International Publishing, 2020, pp. 572--582.

\bibitem{carlini2017towards}
N.~Carlini and D.~Wagner, ``Towards evaluating the robustness of neural
  networks,'' in \emph{2017 IEEE Symposium on Security and Privacy (SP)}.\hskip
  1em plus 0.5em minus 0.4em\relax IEEE, 2017, pp. 39--57.

\bibitem{madry2018towards}
A.~Madry, A.~Makelov, L.~Schmidt, D.~Tsipras, and A.~Vladu, ``Towards deep
  learning models resistant to adversarial attacks,'' in \emph{International
  Conference on Learning Representations}, 2018.

\bibitem{papernot2016distillation}
N.~Papernot, P.~McDaniel, X.~Wu, S.~Jha, and A.~Swami, ``Distillation as a
  defense to adversarial perturbations against deep neural networks,'' in
  \emph{2016 IEEE Symposium on Security and Privacy (SP)}.\hskip 1em plus 0.5em
  minus 0.4em\relax IEEE, 2016, pp. 582--597.

\bibitem{xu2017feature}
W.~Xu, D.~Evans, and Y.~Qi, ``Feature squeezing: Detecting adversarial examples
  in deep neural networks,'' \emph{arXiv preprint arXiv:1704.01155}, 2017.

\bibitem{samangouei2018defense}
P.~Samangouei, M.~Kabkab, and R.~Chellappa, ``{Defense-GAN}: Protecting
  classifiers against adversarial attacks using generative models,'' in
  \emph{International Conference on Learning Representations}, 2018.

\bibitem{liao2018defense}
F.~Liao, M.~Liang, Y.~Dong, T.~Pang, X.~Hu, and J.~Zhu, ``Defense against
  adversarial attacks using high-level representation guided denoiser,'' in
  \emph{Proceedings of the IEEE Conference on Computer Vision and Pattern
  Recognition}, 2018, pp. 1778--1787.

\bibitem{liu2022survey}
S.~Liu, H.~Zhang, and Y.~Jin, ``A survey on surrogate-assisted efficient neural
  architecture search,'' \emph{arXiv:2206.01520}, 2022.

\bibitem{tan2019mnasnet}
M.~Tan, B.~Chen, R.~Pang, V.~Vasudevan, M.~Sandler, A.~Howard, and Q.~V. Le,
  ``{MnasNet}: Platform-aware neural architecture search for mobile,'' in
  \emph{Proceedings of the IEEE Conference on Computer Vision and Pattern
  Recognition}, 2019, pp. 2820--2828.

\bibitem{hsu2018monas}
C.-H. Hsu, S.-H. Chang, J.-H. Liang, H.-P. Chou, C.-H. Liu, S.-C. Chang, J.-Y.
  Pan, Y.-T. Chen, W.~Wei, and D.-C. Juan, ``{MONAS}: Multi-objective neural
  architecture search using reinforcement learning,'' \emph{arXiv preprint
  arXiv:1806.10332}, 2018.

\bibitem{dai2019chamnet}
X.~Dai, P.~Zhang, B.~Wu, H.~Yin, F.~Sun, Y.~Wang, M.~Dukhan, Y.~Hu, Y.~Wu,
  Y.~Jia \emph{et~al.}, ``Chamnet: Towards efficient network design through
  platform-aware model adaptation,'' in \emph{Proceedings of the IEEE/CVF
  Conference on Computer Vision and Pattern Recognition}, 2019, pp.
  11\,398--11\,407.

\bibitem{elsken2018efficient}
T.~Elsken, J.~H. Metzen, and F.~Hutter, ``Efficient multi-objective neural
  architecture search via lamarckian evolution,'' \emph{arXiv preprint
  arXiv:1804.09081}, 2018.

\bibitem{yang2020cars}
Z.~Yang, Y.~Wang, X.~Chen, B.~Shi, C.~Xu, C.~Xu, Q.~Tian, and C.~Xu, ``{CARS}:
  Continuous evolution for efficient neural architecture search,'' in
  \emph{Proceedings of the IEEE/CVF Conference on Computer Vision and Pattern
  Recognition}, 2020, pp. 1829--1838.

\bibitem{zhu2019multi}
H.~Zhu and Y.~Jin, ``Multi-objective evolutionary federated learning,''
  \emph{IEEE Transactions on Neural Networks and Learning Systems}, vol.~31,
  no.~4, pp. 1310--1322, 2020.

\bibitem{zhu2021real}
------, ``Real-time federated evolutionary neural architecture search,''
  \emph{IEEE Transactions on Evolutionary Computation}, vol.~26, no.~2, pp.
  364--378, 2021.

\bibitem{hu2021accelerating}
S.~Hu, R.~Cheng, C.~He, Z.~Lu, J.~Wang, and M.~Zhang, ``Accelerating
  multi-objective neural architecture search by random-weight evaluation,''
  \emph{Complex \& Intelligent Systems}, pp. 1--10, 2021.

\bibitem{lu2020nsganetv2}
Z.~Lu, K.~Deb, E.~Goodman, W.~Banzhaf, and V.~N. Boddeti, ``Nsganetv2:
  Evolutionary multi-objective surrogate-assisted neural architecture search,''
  in \emph{European Conference on Computer Vision}.\hskip 1em plus 0.5em minus
  0.4em\relax Springer, 2020, pp. 35--51.

\bibitem{deb2002fast}
K.~Deb, A.~Pratap, S.~Agarwal, and T.~Meyarivan, ``A fast and elitist
  multiobjective genetic algorithm: {NSGA-II},'' \emph{IEEE Transactions on
  Evolutionary Computation}, vol.~6, no.~2, pp. 182--197, 2002.

\bibitem{ning2020a}
X.~Ning, Y.~Zheng, T.~Zhao, Y.~Wang, and H.~Yang, ``A generic graph-based
  neural architecture encoding scheme for predictor-based nas,'' in
  \emph{Computer Vision -- ECCV 2020}, A.~Vedaldi, H.~Bischof, T.~Brox, and
  J.-M. Frahm, Eds.\hskip 1em plus 0.5em minus 0.4em\relax Cham: Springer
  International Publishing, 2020, pp. 189--204.

\bibitem{jensen2004helper}
M.~T. Jensen, ``Helper-objectives: Using multi-objective evolutionary
  algorithms for single-objective optimisation,'' \emph{Journal of Mathematical
  Modelling and Algorithms}, vol.~3, no.~4, pp. 323--347, 2004.

\bibitem{stein1987large}
M.~Stein, ``Large sample properties of simulations using latin hypercube
  sampling,'' \emph{Technometrics}, vol.~29, no.~2, pp. 143--151, 1987.

\bibitem{wang2020transfer}
H.~Wang, Y.~Jin, C.~Yang, and L.~Jiao, ``Transfer stacking from low-to
  high-fidelity: A surrogate-assisted bi-fidelity evolutionary algorithm,''
  \emph{Applied Soft Computing}, vol.~92, p. 106276, 2020.

\bibitem{deb1995simulated}
K.~Deb, R.~B. Agrawal \emph{et~al.}, ``Simulated binary crossover for
  continuous search space,'' \emph{Complex Systems}, vol.~9, no.~2, pp.
  115--148, 1995.

\bibitem{krizhevsky2010cifar}
A.~Krizhevsky, V.~Nair, and G.~Hinton, ``Cifar-10 (canadian institute for
  advanced research),'' \emph{URL http://www. cs. toronto. edu/kriz/cifar.
  html}, vol.~8, 2010.

\bibitem{krizhevsky2009learning}
A.~Krizhevsky and G.~Hinton, ``Learning multiple layers of features from tiny
  images,'' University of Toronto, Toronto, Ontario, Tech. Rep., 2009.

\bibitem{netzer2011reading}
Y.~Netzer, T.~Wang, A.~Coates, A.~Bissacco, B.~Wu, and A.~Y. Ng, ``Reading
  digits in natural images with unsupervised feature learning,'' in \emph{NIPS
  Workshop on Deep Learning and Unsupervised Feature Learning 2011}, 2011.

\bibitem{sandler2018mobilenetv2}
M.~Sandler, A.~Howard, M.~Zhu, A.~Zhmoginov, and L.-C. Chen, ``Mobilenetv2:
  Inverted residuals and linear bottlenecks,'' in \emph{Proceedings of the IEEE
  Conference on Computer Vision and Pattern Recognition}, 2018, pp. 4510--4520.

\bibitem{zhou2021two}
Q.~Zhou, J.~Wu, T.~Xue, and P.~Jin, ``A two-stage adaptive multi-fidelity
  surrogate model-assisted multi-objective genetic algorithm for
  computationally expensive problems,'' \emph{Engineering with Computers},
  vol.~37, no.~1, pp. 623--639, 2021.

\end{thebibliography}
\end{document}